\documentclass{article}

\usepackage{arxiv}

\usepackage[utf8]{inputenc} 
\usepackage[T1]{fontenc}    

\usepackage{graphicx}

\usepackage{bookmark}

\usepackage{graphicx}
\usepackage{amsmath}
\usepackage{amsthm}
\usepackage{amssymb}

\usepackage{xcolor}
\newtheorem{myDef}{Definition}

\usepackage{algorithm}
\usepackage{algorithmic}
\usepackage{appendix}

\usepackage{times}
\usepackage{soul}
\usepackage{url}

\usepackage[utf8]{inputenc}
\usepackage[small]{caption}
\usepackage{booktabs}
\usepackage{multirow}
\usepackage{hyperref}

\title{Boosting-based Construction of BDDs for Linear Threshold Functions and 
Its Application to Verification of Neural Networks}

\author{
	Yiping Tang\\
	Kyushu University\\
	\texttt{tang.yiping.641@s.kyushu-u.ac.jp} \\
	\And
	\href{https://orcid.org/0000-0000-0000-0000}{\includegraphics[scale=0.06]{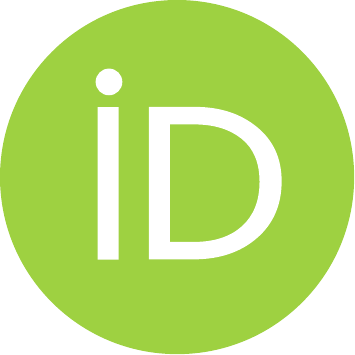}
		\hspace{1mm}Kohei Hatano}\\
	Kyushu University/RIKEN AIP\\
	\texttt{hatano@inf.kyushu-u.ac.jp} \\
	\And
	\href{https://orcid.org/0000-0000-0000-0000}{\includegraphics[scale=0.06]{orcid.pdf}
		\hspace{1mm}Eiji Takimoto}\\
	Kyushu University\\
	\texttt{eiji@inf.kyushu-u.ac.jp} \\
}

\renewcommand{\shorttitle}{Boosting-based Construction of BDDs for LTF and 
Its Application to Verification of NNs}

\hypersetup{
pdftitle={Boosting-based Construction of BDDs for Linear Threshold Functions and 
Its Application to Verification of Neural Networks},
pdfsubject={},
pdfauthor={},
pdfkeywords={},
}

\newcommand{\our}{ABDD}

\newcommand{\calH}{\mathcal{H}}
\newcommand{\defeq}{\triangleq}
\newcommand{\calI}{\mathcal{I}}

\newtheorem{prop}{Proposition}

\newtheorem{theorem}{Theorem}
\newtheorem{corollary}{Corollary}
\newtheorem{lemma}[theorem]{Lemma}

\begin{document}
\maketitle

\begin{abstract}

Understanding the characteristics of neural networks is important but difficult 
due to their complex structures and behaviors.
Some previous work proposes to transform neural networks into equivalent Boolean expressions 
and apply verification techniques for characteristics of interest. This approach is promising since rich results of verification techniques 
for circuits and other Boolean expressions can be readily applied. The bottleneck is the time complexity of the transformation.
More precisely, (i) each neuron of the network, i.e., a linear threshold function, 
is converted to a Binary Decision Diagram (BDD), 
and (ii) they are further combined into some final form, such as Boolean circuits.
For a linear threshold function with $n$ variables, an existing method takes $O(n2^{\frac{n}{2}})$ time to construct an ordered BDD 
of size $O(2^{\frac{n}{2}})$ consistent with some variable ordering.
However, it is non-trivial to choose a variable ordering producing a small BDD among $n!$ candidates.

We propose a method to convert a linear threshold function to a specific form of a BDD based on the boosting approach in the machine learning literature. 
Our method takes $O(2^n \text{poly}(1/\rho))$ time and outputs BDD of size $O(\frac{n^2}{\rho^4}\ln{\frac{1}{\rho}})$,
where $\rho$ is the margin of some consistent linear threshold function. Our method does not need to search for good variable orderings and 
produces a smaller expression when the margin of the linear threshold function is large. 
More precisely, our method is based on our new boosting algorithm, which is of independent interest.
We also propose a method to combine them into the final Boolean expression representing the neural network.
In our experiments on verification tasks of neural networks, 
our methods produce smaller final Boolean expressions, 
on which the verification tasks are done more efficiently. 
	
\end{abstract}

\keywords{Convolutional Neural Network  \and Binary decision diagram \and Boosting \and Verification.}

\section{Introduction}

Interpretability of Neural Networks (NNs) has been relevant 
since their behaviors are complex to understand. 
Among many approaches to improve interpretability, 
some results apply verification techniques of Boolean functions to understand NNs, 
where NNs are represented as an equivalent Boolean function and 
then various verification methods are used to check criteria such as robustness
~\cite{LiuMXK20,MangalNO19,WengZCYSGHD18,YuQLZWC19,ZhengSLG16}. 
This approach is promising 
in that rich results of Boolean function verification can be readily applied.
The bottleneck, however, is to transform a NN into some representation of 
the equivalent Boolean function.

A structured way of transforming NNs to Boolean function representations is proposed by~\cite{ShiSDC20}.
They proposed (i) to transform each neuron, i.e., a linear threshold function, 
into a Binary Decision Diagram (BDD) 
and then (ii) to combine BDDs into a final Boolean function representations 
such as Boolean circuits. 
In particular, the bottleneck is the transformation of a linear threshold function to a BDD. 
To do this, they use the transformation method of~\cite{ChanD03}. 
The method is based on dynamic programming, and its time complexity is $O(n2^{\frac{n}{2}})$ and 
the size of resulting BDD is $O(2^{\frac{n}{2}})$, where $n$ is the number of the variables.
In addition, the method requires a fixed order of $n$ variables as an input 
and outputs the minimum BDD consistent with the order. 
Thus, to obtain the minimum BDD, it takes $O(n!n2^{\frac{n}{2}})$ time by examining $n!$ possible orderings.
Even if we avoid the exhaustive search of orderings, it is non-trivial to choose a good ordering.

In this paper, 
we propose an alternative method to obtain a specific form of BDD representation (named Aligned Binary Decision Diagram, ABDD) of a linear threshold function. 
Our approach is based on \emph{Boosting}, a framework of machine learning 
which combines base classifiers into a better one. 
More precisely, 
our method is a modification of the boosting algorithm of Mansour and McAllester~\cite{mansour2002}.
Given a set of labeled instances of a linear threshold function, 
their algorithm constructs a BDD that is consistent with the instances in a top-down greedy way.
The algorithm can be viewed as a combination of greedy decision tree learning and a process of merging nodes.
Given a linear threshold function $f(x)=\sigma(w\cdot x +b)$ where $\sigma$ is the step function, 
we can apply the algorithm of Mansour and McAllester by feeding all $2^n$ possible labeled instances of $f$ and 
obtain a BDD representation of $f$ of the size $O(\frac{n^2}{\rho^4}\ln{\frac{1}{\rho}})$ 
in time $O(2^n \text{poly}(1/\rho))$, 
where $\rho$ is the margin of $f$, defined as $\rho=\min_{x \in \{-1,1\}^n}|w\cdot x +b|/\|w\|_1$.
An advantage of the method is that the resulting BDD is small if the linear threshold function has a large margin.
Another merit is that the method does not require a variable ordering as an input.
However, in our initial investigation, 
we observe that the algorithm is not efficient enough in practice. 
Our algorithm, in fact, a boosting algorithm, is obtained by modifying their algorithm so that 
we only use one variable (base classifier) in each layer. 
We show that our modification still inherits the same theoretical guarantees as Mansour and McAllester's.
Furthermore, surprisingly, the small change makes the merging process more effective and produces much smaller BDDs in practice.
Our modification might look easy but is non-trivial in a theoretical sense. 
To achieve the same theoretical guarantee, we introduce a new information-theoretic criterion to choose variables that is different from the previous work. That is one of our technical contributions.
\begin{table*}[]
\centering
\caption{Time and size for several methods to convert to DDs from a given linear threshold function (LTF, for short) of margin $\rho$. 
The fourth result is only for LTFs with integer weights whose $L2$-norm is $W$.}
\begin{tabular}{|c|c|c|c|}
\hline
Method  & DD type & Size& Time  \\ \hline
\cite{ChanD03}   & OBDD   & $O(2^{\frac{n}{2}})$ &$O(n! n2^{\frac{n}{2}})$ \\ \hline
\cite{mansour2002} & BDD & $O(\frac{n^2}{\rho^4}\ln{\frac{1}{\rho}})$ & $O(2^n \text{poly}(1/\rho))$\\ \hline
Ours & \our & $O(\frac{n^2}{\rho^4}\ln{\frac{1}{\rho}})$& $O(2^n \text{poly}(1/\rho))$ \\ \hline
\hline
(cf.~\cite{ShiSDC20}) & OBDD & $O(nW)$& $O(nW)$ \\ \hline
\end{tabular}
\label{tab1}
\end{table*}

In our experiments on verification tasks of Convolutional Neural Networks (CNNs), by following the same procedures as~\cite{ShiSDC20},
we construct smaller BDDs and resulting Boolean representations of CNNs faster than in previous work, thus contributing to more efficient verification.

This paper is organized as follows: Section 2 overviews the preliminaries of binary NN (BNN), BDD, Ordered BDD (OBDD), and Aligned BDD (ABDD). Section 3 and 4 detail our proposed method to construct \our. Section 5 details the construction of the Boolean circuit and SDD. Section 6 handles the experimental results with analysis, followed by the conclusion in Section 7. 
\paragraph{Related work}
The work in~\cite{narodytska2018} proposed a precise Boolean encoding of BNNs that allows easy network verification. However, it only works with small-sized networks.~\cite{shih2019} leveraged the Angluin-style learning algorithm to convert the BNN (the weights and input are binarized as $\{-1,1\}$) and OBDD into Conjunctive Normal Form (CNF) and then used the Boolean Satisfiability (SAT) solver to verify the equivalence of its produced CNF. However, they modified the OBDD several times and utilized limited binary network weights. 
~\cite{ChanD03} suggested a method to convert a linear threshold classifier with real-valued weight into OBDD. However, their approach owns time complexity of $O(n!n2^{\frac{n}{2}})$ and OBDD size complexity of $O(2^{\frac{n}{2}})$ via searching the full ordering, which increases exponentially when $n$ becomes larger. 
Still, ~\cite{narodytska2018,shih2019,ChanD03} can only handle small dimension NN weight, and the large Boolean expression was represented as Sentential Decision Diagram (SDD), which owns enormous time complexity. Moreover,~\cite{ChorowskiZ11} proposed a rule extraction method inspired by the 'rule extraction as learning' approach to express NN into a Reduced Ordered DD (RODD), which has the time complexity of $O(n2^{2n})$.\\

\section{Preliminaries}

\subsection{Binary Neural Network}
A binary neural network (BNN) is a variant of the standard NN with binary inputs and outputs~\cite{BshoutyTW98}. 
In this paper, each neural unit, with a step activation function $\sigma$, is formulated as follows:
\begin{equation}\label{bnnunit}
\begin{split}
\sigma(\sum_i x_i w_i + b)=
\begin{cases} 
1, &  \sum_i x_i w_i + b \geq 0\\
-1, & \text{otherwise}
\end{cases}
\end{split}
\end{equation} 
where $x\in \{-1,1\}^n$, $w\in \mathbb{R}^n$ and $b\in \mathbb{R}$ are the input, 
the weight vector and the bias of this neural unit, respectively. 

\subsection{Definition of BDD, OBDD and ABDD}
A binary decision diagram (BDD) $T$ is defined as a tuple $T=(V,E,l)$ with the following properties: (1) $(V,E)$ is a directed acyclic graph with a root and two leaves, 
where $V$ is the set of nodes, $E$ is the set of edges such that $E=E_- \cup E_+$, $E_-\cap E_+ =\varnothing$. Elements of $E_+$ and $E_-$'s are called $+$-edges, 
and $-$-edges, respectively. Let $L=\{0$-leaf$,1$-leaf$\}\subset V$ be the set of leaves.
For each $v\in V$, there are two child nodes $v^-,v^+ \in V$ such that $(v,v^-)\in E_-$ and $(v,v^+)\in E_+$.
(2) $l$ is a function from $V\setminus L$ to $[n]$.

Given an instance $x\in {\{-1,1\}}^n$ and a BDD $T$, we define the corresponding path 
$P(x)=(v_0,v_1,\ldots,v_{k-1},v_{k})\in V^*$ over $T$ from the root to a leaf as follows:
(1) $v_0$ is the root. (2) for any $j=0,\ldots k-1$, we have 
$(v_{j},v_{j+1})\in E_+ \Leftrightarrow x_{l(v_{j})}=1$ and 
$(v_{j},v_{j+1})\in E_- \Leftrightarrow x_{l(v_{j})}=-1$.
(3) $v_k$ is a leaf node. 
We say that an instance $x\in \{-1,1\}^n$ reaches node $u$ in $T$, if $P(x)$ contains $u$. 
Then, a BDD $T$ naturally defines the following function $h_T:{\{-1,1\}}^n\rightarrow \{-1,1\}$ such that
\begin{equation}\label{eq000}
\begin{split}
h_T(x)\triangleq
\begin{cases} 
-1,  &  \text{$x$ reaches } 0\mbox{-leaf} \\
1, &  \text{$x$ reaches } 1\mbox{-leaf}.
\end{cases}
\end{split}
\end{equation}
Given a BDD $T=(V,E,l)$, we define the depth of a node $u\in V$ as the length of the longest path from the root to $u$.  
An ordered BDD (OBDD) $T=(V,E,l)$ is a BDD satisfying an additional property: There is a strict total order $<_{[n]}$ on $[n]$ such that for any path $P=(v_0,\ldots,v_k)$ on from the root to a leaf, and any nodes $v_i$ and $v_j$ $(i< j < k)$, $l(v_i) <_{[n]} l(v_j)$.
An Aligned BDD (ABDD) $T=(V,E,l)$ is defined as a BDD satisfying that for any nodes $u,v\in V\setminus L$ with the same depth, $l(u)=l(v)$.
We employ $v_{i,j}$ to appear the positional information of a node in the BDD, where $j$ represents the depth of the node, and $i$ represents the position of the node at depth $j$.

\begin{figure}[!http]
\begin{center}
\begin{minipage}[t]{0.3\linewidth}
\centering
\includegraphics[scale=0.275]{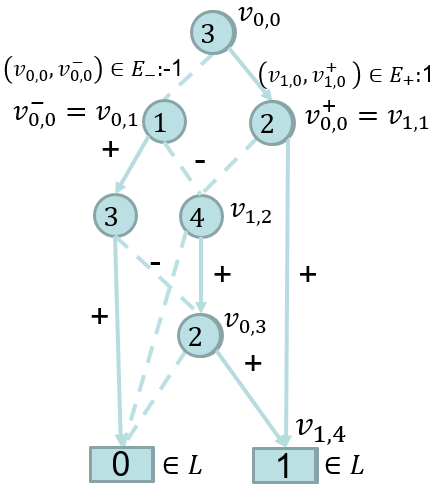} 
\\
{A}
\end{minipage}
\begin{minipage}[t]{0.3\linewidth}
\centering
\includegraphics[scale=0.275]{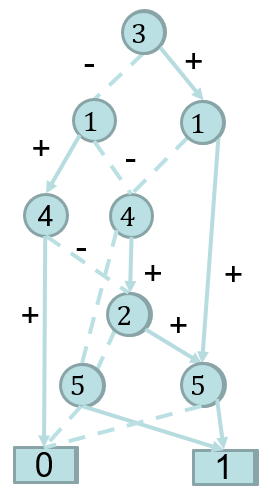} 
\\
{B}
\end{minipage}
\begin{minipage}[t]{0.3\linewidth}
\centering
\includegraphics[scale=0.275]{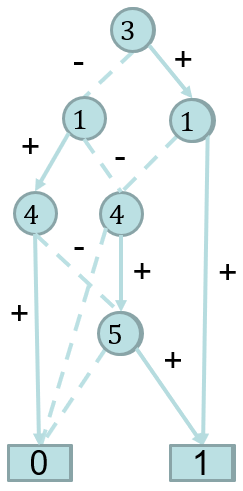} 
\\
{C}
\end{minipage}
\caption{Examples of BDD (A), OBDD (B) and ABDD (C).
To express the same linear threshold function in BDD form: nodes at the same depth can be labeled by different variables, which means that the number of variables does not limit the depth of BDD; in OBDD form: the nodes at the same depth are all labeled by the same variables, which results in the depth of OBDD is smaller than the number of variables; in \our{} form: nodes at the same depth are labeled by the same variables, and the depth of \our{} only depends on the reduction of entropy to 0 in our algorithm.}
\label{dds}
\end{center}
\end{figure}

BDD, OBDD and ABDD are illustrated in Figure~\ref{dds}.


\subsection{Instance-based Robustness (IR), Model-based Robustness (MR) and Sample-based Robustness (SR)}

Robustness is a fundamental property of the neural network, 
which represents the tolerance of the network to noise or white attacks. 
For binary input images, the robustness $k$ represents that as long as at least $k$ 
pixels are flipped from 0 to 1 or 1 to 0, and the neural network's output will be changed.

We define the IR and MR of a network as follows.

\begin{myDef}{(Instance-based Robustness)~\cite{ShiSDC20}}\label{Instance-b}\\ 
Consider a classification function $f:{\{-1,1\}}^n \rightarrow \{-1,1\}$ and a given instance $x$.
The robustness of the classification of $x$ by $f$, denoted by $r_f(x)$.
If $f$ is not a trivial function (always $True$ or $False$), 
\begin{equation}\label{instance-base}
\begin{split}
r_f(x)=\min_{x^\prime:f(x)\neq f(x^\prime)} dis(x,x^\prime)
\end{split}
\end{equation}
where $dis(x,x^\prime)$ denotes the Hamming distance between $x$ and $x^\prime$.
\end{myDef}

\begin{myDef}{(Model-based Robustness)~\cite{ShiSDC20}}\label{model-d} \\
Consider a classification function $f:{\{-1,1\}}^n\rightarrow \{-1,1\}$.
The Model-based Robustness of $f$ is defined as:
\begin{equation}\label{model-base}
\begin{split}
MR(f)=\frac{1}{2^n}\sum_x r_f(x).
\end{split}
\end{equation}
\end{myDef}

However, we consider that computing MR on full-size data is not practically meaningful. 
In practical applications, robustness validation based on sample data is common. 
Here, we regard the samples in the dataset as instances randomly extracted from the full-size data under the uniform distribution.
Then, we have the following definition.

\begin{myDef}{(Sample-based Robustness, SR)}\label{sr} \\
Consider a classification function $f:{\{-1,1\}}^n\rightarrow \{-1,1\}$.
Given a sample $S$ under uniform distribution from $\{-1,1\}^n$.
The Sample-based Robustness of $f$ is defined as:
\begin{equation}\label{sample-base}
\begin{split}
SR(f)=\frac{1}{|S|}\sum_{x\in S} r_f(x).
\end{split}
\end{equation}
\end{myDef}

\subsection{Overview of our method}
In Section 3, 
we propose an algorithm that constructs an ABDD whose training error is small with respect to a given sample of some target Boolean function. 
Our algorithm is based on boosting, which is an effective approach in machine learning 
that constructs a more accurate classifier by combining ``slightly accurate'' classifiers. 
In Section 4, 
we apply our boosting algorithm for finding an equivalent ABDD with a given linear threshold function.
Under a natural assumption that the linear threshold function has a large ``margin'', 
we show the size of the resulting ABDD is small. 
In Section 5, 
we show how to convert a given BNN to an equivalent Boolean expression suitable for verification tasks.
More precisely, 
(i) Each neural unit is converted to an equivalent ABDD by applying our boosting algorithm, 
(ii) each ABDD is further converted to a Boolean circuit, and 
(iii) all circuits are combined into the final circuit, which is equivalent to the given BNN.
Furthermore, for a particular verification task, we convert the final circuit to 
an equivalent sentential decision diagram (SDD).

\section{Boosting}
\subsection{Problem Setting}
Boosting is an approach to constructing a strongly accurate classifier 
by combining weakly accurate classifiers. 
We assume some unknown target function $f:\{-1,1\}^n \to \{-1,1\}$. 
Given a sample 
$S=((x_1,f(x_1)),\dots,(x_m,f(x_m))) \in (\{-1,1\}^n\times \{-1,1\})^m$
of $m$ instances labeled by $f$ and a precision parameter $\varepsilon$, 
we want to find a classifier $g:\{-1,1\}^n \to \{-1,1\}$ such that 
its training error $\Pr_U\{g(x) \neq f(x)\}\leq \varepsilon$, 
where $U$ is the uniform distribution over $S$. 
We are also given a set $\calH$ of base classifiers from $\{-1,1\}^n$ to $\{-1,1\}$. 
We assume the following assumption which is standard in the boosting literature~\cite{mansour2002}.  

%

\begin{myDef}{(Weak Hypotheses Assumption (WHA))}\label{wha}\\
A hypothesis set $\calH$ satisfies $\gamma$-Weak Hypothesis Assumption (WHA) for the target function 
$f:\{-1,1\}^n \to \{-1,1\}$ 
if 
for any distribution $d$ over $\{-1,1\}^n$, there exists $h\in \calH$ such that 
$edge_{d,f}(h)\triangleq\sum_{x\in \{-1,1\}^n} d_x f(x) h(x)\geq \gamma$. 
\end{myDef} 
Intuitively, WHA ensures the set of $\calH$ of hypotheses and $f$ are ``weakly'' related to each other. 
The edge function $edge_{d,f}(h)$ takes values in $[-1,1]$ and equals to $1$ if $f=h$. 
Under WHA, we combine hypotheses of $\calH$ into a final hypothesis $h_T$ represented by an ABDD $T$.

Our analysis is based on a generalized version of entropy~\cite{kearns1999}. 
A pseudo-entropy $G:[0,1] \to [0,1]$ is defined as $G(q)\defeq2\sqrt{q(1-q)}$. 
Like the Shannon entropy, $G$ is concave and $G(1/2)=1$, and $G(0)=G(1)=0$. 
In particular, $\min(q, 1-q)\leq G(q)$. 
Then we will introduce the conditional entropy of $f$ given a sample $S$ and an ABDD $T$. 
For each node $u$ in $T$, let $p_u \defeq\Pr_U\{\text{$x$ reaches $u$}\}$ and 
$q_u \defeq\Pr_U\{f(x)=1 \mid \text{$x$ reaches $u$}\}$, respectively. 
Let $N(T)$ be the set of nodes in $V\setminus L$ whose depth is the maximum. 
We further assume that for each node $u$ in $N(T)$, the $x$ in $u$ is assigned to $1$-leaf if $q_u \geq 1/2$, and is assigned to $0$-leaf, otherwise.
Then, observe that $\Pr_U\{f(x)\neq h_T(x)\}=\sum_{u \in N(T)}p_u \min (q_u, 1-q_u)$.  
The conditional entropy of $f$ given an ABDD $T$ with respect to the distribution $U$ is defined as 
\begin{equation}\label{eq1}
    \begin{split}
    H_U(f|T)=\sum_{u\in N(T)} p_u G(q_u).
    \end{split}
\end{equation} 
Then the conditional entropy gives an upper bound of the training error as follows.

\begin{prop}\label{propp0}
$\Pr_U\{h_T(x)\neq f(x)\}\leqslant H_U(f|T)$.
\end{prop}

Therefore, it is sufficient to find an ABDD $T$ 
whose conditional entropy $H_U(f|T)$ is less than $\varepsilon$.

We will further use the following notations and definitions. 
Given an ABDD $T$, $S$ and $u \in N(T)$, let $S_u=\{(x,y)\in S \mid \text{$x$ reaches $u$}\}$.
The entropy $H_d(f)$ of $f:\{-1,1\}^n \to \{-1,1\}$ 
with respect to a distribution $d$ over $\{-1,1\}^n$
is defined as $H_d(f)\defeq G(q)$, where $q=\Pr_d\{f(x)=1\}$. 
The conditional entropy $H_d(f|h)$ of $f$ given $h:\{-1,1\}^n \to \{-1,1\}$ with respect to $d$ is defined as 
$H_d(f|h)=\Pr_d\{h(x)=1\}G(q^+) + \Pr_d\{h(x)=-1\}G(q^-)$, 
where $q^{\pm}=\Pr_d\{f(x)=1 \mid h(x)=\pm1\}$, repectively.
\subsection{Our Boosting Algorithm}
Our algorithm is a modification of the boosting algorithm 
proposed by Mansour and McAllester~\cite{mansour2002}.
Both algorithms learn Boolean functions in the form of BDDs in a top-down manner.
The difference between our algorithm and Mansour and McAllester's algorithm lies in the construction of the final Boolean function, where ours utilizes ABDDs, while Mansour and McAllester's algorithm does not. 
Although this change may appear subtle, it necessitates a new criterion for selecting hypotheses in $\calH$ and demonstrates improved results in our experiments.

\begin{figure*}[h]
\begin{center}
\includegraphics[scale=0.18]{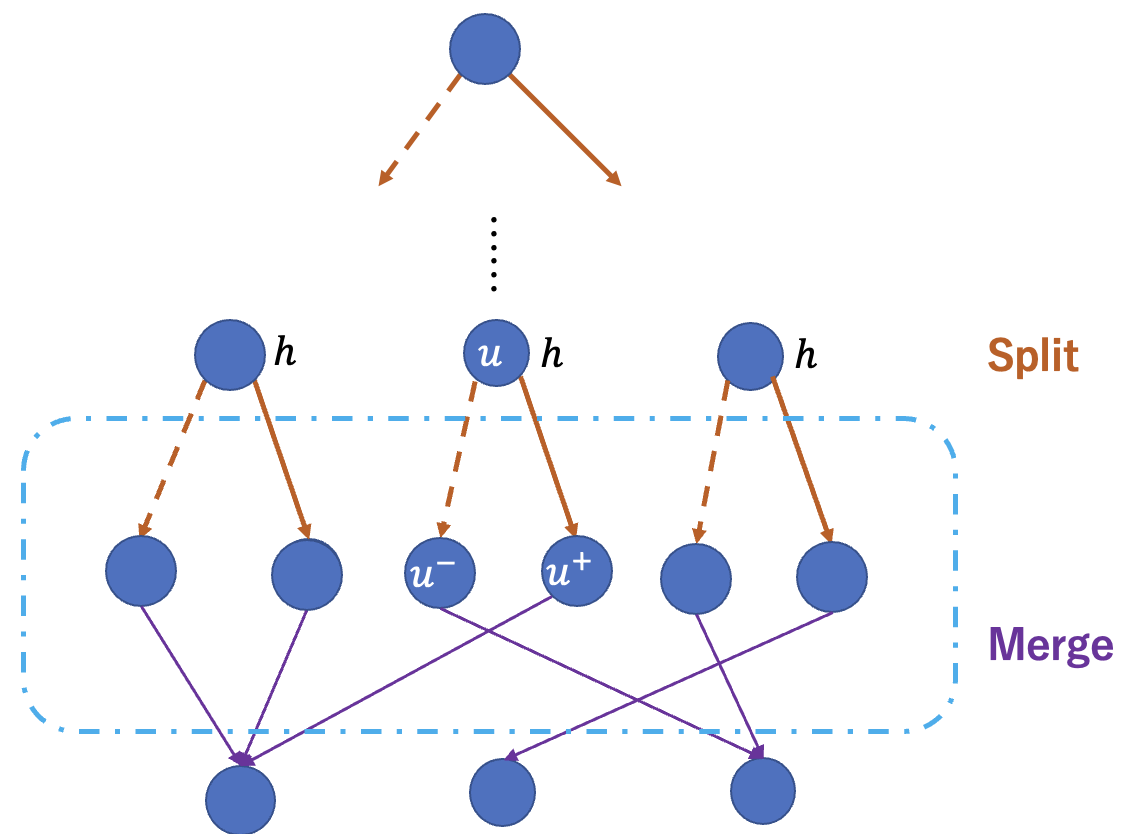} 
\caption{Illustration of our boosting algorithm. 
The blue dotted line part represents the process of merging the split temporary nodes into new nodes by searching the equivalence space
that does not appear in the \our. 
Following algorithm~\ref{alg1}, we find some hypothesis $h$ to construct child nodes in the new layer
and then merge nodes afterward.}\label{hiddenl}
\end{center}
\end{figure*}

Our boosting algorithm iteratively grows an ABDD by adding a new layer at the bottom. 
More precisely, at each iteration $k$, given the current ABDD $T_k$, 
the algorithm performs the following two consecutive processes 
(as illustrated in Figure~\ref{hiddenl}).
\begin{description}
    \item[Split:] 
    It chooses a hypothesis $h_k\in \calH$ using some criterion and 
adds two child nodes for each node in $N(T_k)$ in the next layer, 
where each child corresponds to $\pm 1$ values of $h_k$.
Let $T_k'$ be the resulting DD. 
\item[Merge:]
It merges nodes in $N(T_k')$ according to some rule 
and let $T_{k+1}$ be the ABDD after the merge process.
\end{description}

The full description of the algorithm is given in Algorithm~\ref{alg1} and~\ref{algsplit}, 
respectively.
For the split process, it chooses the hypothesis $h_k$ 
maximizing the edge $edge_{\hat{d},f}(h)$ with respect to the distribution $\hat{d}$ specified in 
(\ref{balance_dist}). 
For the merge process, we use the same way in the algorithm of Mansour and McAllester~\cite{mansour2002}.

\begin{myDef}{~\cite{mansour2002}}\label{mydef1} 
For $\delta$ and $\lambda$ ($0<\delta,\lambda <1$),
a $(\delta,\lambda)$-net $\calI$ is defined as 
a set of intervals $[v_0,v_1],[v_1,v_2],\dots,[v_{w-1},v_w]$ such that 
(i) $v_0=0$, $v_w=1$, (ii) for any $I_k=[v_{k-1},v_k]$ and $q \in I_k$, 
$\max_{q' \in I_k}G(q')\leq \max\{\delta, (1+\lambda)G(q) \}$. 
\end{myDef}
Mansour and McAllester showed a simple construction of $(\delta,\lambda)$-net 
with length $w=O((1/\lambda)\ln (1/\delta))$~\cite{mansour2002} and we omit the details.
Our algorithm uses particular $(\delta,\lambda)$-nets for merging nodes.

\subsection{Analyses}


\begin{equation}\label{eq2}
\begin{split}
H_d(Y|T,h)=\sum_{u\in N(T)} (p_{u^+} G(q_{u^+})+p_{u^-} G(q_{u^-}))
\end{split}
\end{equation}
where $p_{u^+}=Pr_d\{x \mbox{ reaches }u | h(x)=1\}$ and $q_{u^+}=Pr_d(f(x)=1|\mbox{$x$ reaches $u$, }\\ h(x)=1)$. $p_{u^-}$ and $q_{u^-}$ are defined similarly.

The following lemmas prove an effective weak learning algorithm 
under a variant balanced distribution $\bar{d}$, and it also can be achieved 
under any distribution. For $(x,y)\in S$:
\begin{equation}\label{vbalancedd}
\begin{split}
\bar{d}_{(x,y)}=
\frac{d_{(x,y)}}{2\sum_{(x^\prime,y) \in S} d_{(x^\prime,y)}}
\end{split}
\end{equation}

\begin{lemma}{\cite{takimoto2003}}\label{lemma1-1}
Let $d$ and $\bar{d}$ be any distribution over $\{-1,1\}^n$ 
and its balanced distribution with respect to some $f:\{-1,1\}^n \to \{-1,1\}$, 
respectively.
Then, for any hypothesis with $edge_{\bar{d},f}(h)\geq \gamma$, 
$H_{d}(f|h)\leqslant(1-\gamma^2/2)H_{d}(f)$.
\end{lemma}


Based on Lemma~\ref{lemma1-1}, we establish the connection between $\gamma$ 
and entropy function, and have a $\gamma^*\in (0,1)$ at each depth to reflect the entropy change
under our algorithm.

\begin{lemma}\label{lemma0} 
Let $\hat{d}$ be the distribution over $S$ specified in (\ref{balance_dist}) 
when $T_k$, $\calH$ and $S$ is given by Algorithm~\ref{algsplit} and let $h_k$ be the output.
If $\calH$ satisfies $\gamma$-WHA, 
then 
the conditional entropy of $f$ with respect to the distribution $U$ over $S$ given $T_k$ and $h_k$ is bounded as 
$H_U(f|h_k,T_k)\leqslant (1-\gamma^{2}/2)H_U(f|T_k)$.
\end{lemma}



\begin{lemma}{(\cite{mansour2002})}\label{lemma-merge} 
    Assume that, before the merge process in Algorithm~\ref{alg1}, 
    $H_U(f|h_k,T_k)\leq (1-\lambda) H_U(f|T_k)$ for some $\lambda$ ($0<\lambda<1$). 
    Then, by merging based on the $(\delta, \eta)$-net with 
    $\delta=(\lambda/6)H_U(f|T_k)$ and $\eta=\lambda/3$, 
    the conditional entropy of $f$ with respect to the distribution $U$ over $S$ given $T_{k+1}$ is bounded as 
    $H_U(f|T_{k+1})\leqslant (1-\lambda/2)H_U(f|T_k)$, 
    where the width of $T_{k+1}$ is $O((1/\lambda)(\ln(1/\lambda)+\ln(1/\varepsilon)))$, 
    provided that $H_U(f|T_k)>\varepsilon$.
    \end{lemma}


\begin{algorithm}[h]
\caption{\our{} Boosting}\label{alg1} 
\hspace*{0.02in} {\bf Input:}
a sample $S\in (\{-1,1\}^n\times ({-1,1}))^m$ of $m$ instances by $f$, 
and a set $\calH$ of of hypotheses, and precision parameter $\varepsilon$ ($0<\varepsilon<1$);\\
\hspace*{0.02in} {\bf Output:}
\our{} $T$;
\begin{algorithmic}[1] 
\STATE initialization: $T_1$ is the ABDD with a root and $0,1$-leaves, $k=1$. 
\REPEAT
\STATE (\textbf{Split}) Let $h_k=Split(T,\calH,S)$ and add child nodes with each node in $N(T_k)$. 
Let $T_k'$ be the resulting ABDD.

\FOR{$u\in N(T_k')$}
\STATE merge $u$ to the $0$-leaf (the $1$ leaf) if $q_u=0$ ($q_u=1$, resp.).
\ENDFOR

\STATE (\textbf{Merge}) Construct a $(\hat{\delta},\hat{\lambda}/3)$-net $\calI_k$ with 
\STATE 
$\hat{\lambda}=1-\frac{H_U(f|T_k,h_k)}{H_U(f|T_k)}$, 
and $\hat{\delta}=\frac{\hat{\lambda} H_U(f|T_k)}{6}$.
\FOR{$I\in \mathcal{I}_k$} 
\STATE merge all nodes $u \in N(T_k')$ such that $q_u \in I$.
\ENDFOR
\STATE Let $T_{k+1}$ be the resulting ABDD and update $k \leftarrow k+1$. 
\UNTIL{$H_U(f|T_k)<\varepsilon$}
\STATE Output $T=T_k$.
\end{algorithmic} 
\end{algorithm}

\begin{algorithm}[h]
\caption{$Split$}\label{algsplit} 
\hspace*{0.02in} {\bf Input:}
ABDD $T$, a set $\calH$ of hypotheses, a set $S$ of $m$ instances labeled by $f$;\\
\hspace*{0.02in} {\bf Output:}
$h \in \calH$;
\begin{algorithmic}[1] 
\FOR{$u\in N(T)$}
\STATE Let \\
$d_{(x,y)}^u=
\begin{small}
\begin{cases} 
\frac{1}{2\times |\{(x^\prime,y^\prime)\in S_u|y=1\}|},  & \mbox{if }y=1  ~\&  ~(x,y)\in S_u\\
\frac{1}{2\times |\{(x^\prime,y^\prime)\in S_u|y=-1\}|}, & \mbox{if }y=-1  ~\&  ~(x,y)\in S_u\\
0, & \mbox{if } (x,y)\notin S_u
\end{cases}
\end{small}$, for $(x,y)\in S$.
\STATE $p^\prime_u =\frac{p_u G(q_u)}{\sum_{u\in \mathbf{L}_k}p_u G(q_u)}$
\STATE Let 
\begin{align}
    \hat{d}_{(x,y)}=\sum_{u\in N(T)}p^\prime_u d^u_{(x,y)} 
    \label{balance_dist}
\end{align}
for $(x,y)\in S$.
\ENDFOR
\STATE Output $h=\arg\max_{h' \in \calH}edge_{\hat{d},f}(h)$.
\end{algorithmic} 
\end{algorithm}
    

Now we are ready to show our main theorem.

\begin{theorem}\label{mainth}
Given a sample $S$ of $m$ instances labeled by $f$, 
and a set $\calH$ of hypotheses satisfying $\gamma$-WHA, 
Algorithm~\ref{alg1} outputs an ABDD $T$ such that 
$\Pr_U\{h_T(x)\neq f(x)\}\leq \varepsilon$. 
The size of $T$ is $O((\ln (1/\varepsilon)/\gamma^4)( \ln (1/\varepsilon) + \ln(1/\gamma)))$ and 
the running time of the algorithm is $\text{poly}(1/\gamma, n)m$.
\end{theorem}   








\section{\our{} Construction}

We now apply the \our{} Boosting algorithm developed in the previous
section to a given linear threshold function $f$ to obtain an
\our{} representation $T$ for $f$.
In particular, we show that the size of $T$ is small when $f$ has a
large margin.

To be more specific, assume that we are given a linear threshold function
$f: \{-1,1\}^n \to \{-1,1\}$ of the form
\[
	f(x) = \sigma(w \cdot x + b)
\]
for some weight vector $w \in \mathbb{R}^n$ and bias $b \in \mathbb{R}$,
where $\sigma$ is the step function, i.e., 
$\sigma(z)$ is $1$ if $z \geq 0$ and $-1$ otherwise.
Note that there are infinitely many $(w,b)$ inducing the same
function $f$. We define the margin $\rho$ of $f$ as the maximum
margin $f(x)(w \cdot x + b)/\|w\|_1$ over all $(w,b)$.
We let our hypothesis set $\calH$ consist of projection functions,
namely, $\calH = \{h_1,h_2,\dots,h_n,h_{n+1}, h_{n+2},\dots,h_{2n}\}$,
where $h_i:x \mapsto x_i$ if $i \leq n$ and $h_i:x \mapsto -x_i$
otherwise, so that we can represent $f$ as
$f(x) = \sigma(\sum_{i=1}^{2n} w_i h_i(x) + b)$ for some
non-negative $2n$-dimensional weight vector $w \geq 0$ and bias $b$.
Then, we can represent the margin $\rho$ of $f$ as the solution of
the optimal solution for the following LP problem:
\begin{align} \label{eq:margin}
& \max_{w,b,\rho} \rho \\
\text{s.t. \quad} &
f(x)(\sum_{i=1}^{2n} w_i h_i(x) + b) \geq \rho
\text{ for any $x \in \{-1,1\}^n$,} \nonumber \\
& w \geq 0, \nonumber \\
& \sum_i w_i = 1. \nonumber
\end{align}

Now we show that $\calH$ actually satisfies $\rho$-WHA for $f$.

\begin{lemma}\label{mainlemma}
Let $f$ be a linear threshold function with margin $\rho$.
Then $\calH$ satisfies $\rho$-WHA.
\end{lemma}

By the lemma and Theorem~\ref{mainth}, we immediately have the following corollary.

\begin{corollary}\label{coroabdd}
Let $f$ be a linear threshold function with margin $\rho$.
Applying the \our{} Boosting algorithm with
the sample
$S = \{(x,f(x)) \mid x \in \{-1,1\}^n\}$
of all ($2^n$) instances,
our hypothesis set $\calH$, and
the precision parameter $\varepsilon = 1/2^n$,
we obtain an \our{} $T$, which is equivalent to $f$
of size $O\left(\frac{n}{\rho^4}(n+\ln(1/\rho))\right)$.
\end{corollary}

\section{Circuit and SDD Construction}
Since we have a method to convert linear threshold functions into a DD representation, the next step is to connect them according to the structure of the NN to form an equivalent $(\vee,\wedge,\neg)$-circuit, which is used to verify SR. Subsequently, we can convert the circuit into an SDD, which is used to verify robustness.

\begin{figure*}[h]
\begin{center}
\centering
\centering
\includegraphics[scale=0.27]{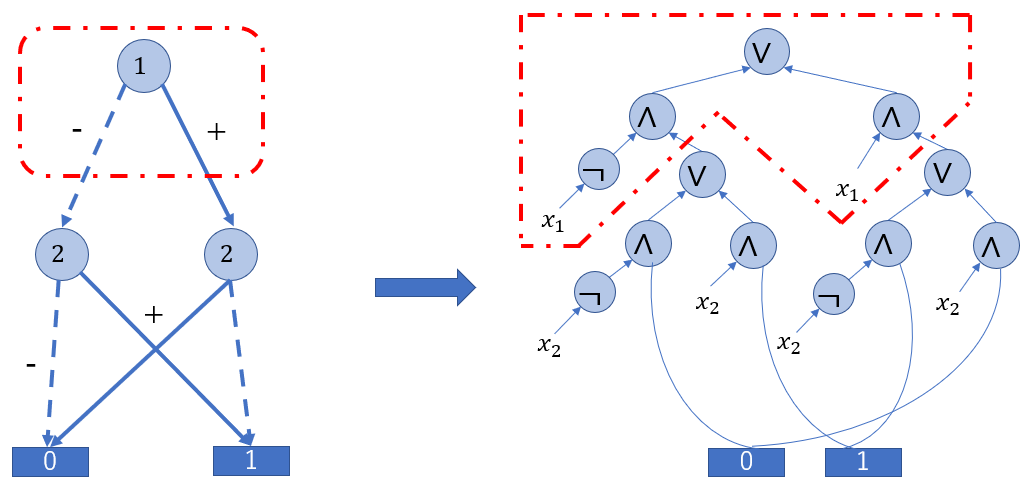} 
\caption{Example of converting BDD/OBDD/\our{} (Left) to the circuit (Right).}
\label{OBDD2circuit}
\end{center}
\end{figure*}

The conversion process from a DD to a circuit is performed in a top-down manner. As shown in Figure~\ref{OBDD2circuit}, the region enclosed by the red dashed box represents a conversion unit. This unit converts a node $x_1$ and its two edges in the DD into four gates and a variable in the circuit. As the next node $x_2$ is connected to a different edge of node $x_1$, we generate a new circuit beginning with an 'or'-gate and connect it to the 'and'-gate obtained from the conversion of an edge associated with $x_1$. This process is repeated until all edges of the nodes reach either the 0-leaf or the 1-leaf. The resulting circuit is equivalent to the DD and its corresponding neural unit.
To construct the NN's equivalent circuit, we utilize Algorithm~\ref{alg1} to generate DDs for each neuron in the NN. These DDs are then converted into circuit form. We combine these equivalent circuits based on the structure of the NN, specifically establishing a one-to-one correspondence between the inputs and outputs of each neuron in the NN and the inputs and outputs of the circuit.
This completes the construction of the NN's equivalent circuit. Such a method is also described in~\cite{ShiSDC20}.

Once we have the equivalent circuit of a neural network (NN), the subsequent step is to convert it into an SDD.
SDD is a subclass of deterministic Decomposable Negation Normal Form (d-DNNF) circuits that assert a stronger decomposability and a more robust form of determinism~\cite{Darwiche11sdd}.
The class of SDDs generalizes that of OBDDs in that, every OBDD can be turned into an SDD in linear time. In contrast, some Boolean functions have polynomial-size SDD representations but only exponential-size OBDD representations~\cite{Bova16}.
In SDD, Boolean functions are represented through the introduction of "decision ($\vee$) nodes" and "conjunction ($\wedge$) nodes."

Indeed, the size complexity of each $apply$ operation between two SDD nodes is proportional to the product of their internal nodes. Consequently, as the complexity of the circuit increases, the construction of the corresponding SDD requires more space.
Considering the conversion process according to \our{}, if the resulting circuit is smaller, it follows that the corresponding SDD constructed from it will also be smaller in size. The size reduction in the circuit conversion directly influences the size of the resulting SDD. Therefore, by optimizing the BDD/circuit representation, we can achieve a more compact SDD.

\section{Experiments}


\subsection{Experimental Setup}
We use the USPS digits dataset of hand-written digits, consisting of $16\times 16$ binary pixel images.
We use the data with labels 0 and 1 for verification of SR.
The NN design we used is similar to~\cite{ShiSDC20}, which has two convolution layers and a full-connected layer.
In training, we use two real-valued-weight convolutional layers (kernel size 3, stride 2; kernel size 2, stride 2) with a sigmoid function and a real-valued-weight fully-connected layer in the NN.
In testing, the sigmoid function is replaced with the step activation function mentioned before.
The experiments are conducted using a CPU of Intel(R) Xeon(R) Gold 2.60GHz. The batch size equals 32 and the utilized learning rate is 0.01 decaying to 10\% at half and three-quarters of all learning epochs. Here, the Stochastic Gradient Descent (SGD) optimizer with a momentum of 0.9 and a weight decay of 0.0001 is used.

\subsection{Sample-based Robustness (SR) validation}

The SR of a standard NN whose output is real-valued is achieved simply by querying the pixels that affect the recognition the most and flipping them until the recognition result changes.
Nevertheless, on the standard methods, for a BNN that has input size $h\times w$, the time cost to compute robustness $k$ is $O({(hw)}^k)$, which is difficult to use in BNNs.

Since BDD, OBDD, and \our{} are easily represented as a circuit, as shown in Figure~\ref{OBDD2circuit}. 
The circuits of several neural units are linked into a circuit $f$ that represents the entire NN according to its network structure. 
Note that we separately verify the SR of the OBDDs generated by the methods based on Theorem 1 and Theorem 2 in~\cite{ShiSDC20} using a circuit representation. The integration of weights and bias in Theorem 2 is performed as follows.

For each $w$ in weight $W$ and bias $b$, 
We set $\alpha=\max\{|w_1|,\ldots,|w_n|,|b|\}$, 
then we turn them to integer weight $\hat{w}_i=\lfloor\frac{10^p}{\alpha} w_i \rfloor $ and bias $\hat{b}=\lfloor\frac{10^p}{\alpha} b \rfloor $ where $p$ is the number of digits of precision.



As for $dis()$ in Definition~\ref{Instance-b}, it's easy to express the $k\leq dis(x,x^\prime)$ between $x$ and $x^\prime$ in a circuit form and likewise convert it to circuit $g_{k,x}$ denoted as follows:
\begin{equation}\label{hamming}
\begin{split}
g_{k,x} (x^\prime)=
\begin{cases} 
1, & if \mid x\oplus x^\prime\mid\leq k\\
-1, & otherwise
\end{cases}
\end{split}
\end{equation} 

We calculate the SR of $f$ on (positive and negative) instance $x$, where have $f(x)=1$ and $f(x)=-1$, by running Algorithm~\ref{instancebaser} of BDD, OBDD and \our{} on 10 CNNs, the results as shown in Table~\ref{tab3211}.
Note that the SR of negative instances can be computed by invoking Algorithm~\ref{instancebaser} on function $\neg f$.

\begin{algorithm}[h]
\caption{$SR$}\label{instancebaser} 
\hspace*{0.02in} {\bf Input:}
circuit $f$, (positive) instance $x\in \{-1,1\}^n$;\\
\hspace*{0.02in} {\bf Output:}
 $r_f(x)$;
\begin{algorithmic}[1] 
\STATE initial: $r_f=0$; 
\FOR{$k=1$ to n}
\IF{$g_{k,x} \land \overline{f}$ is satisfiable}
\STATE break
\ENDIF
\ENDFOR
\RETURN $k$
\end{algorithmic} 
\end{algorithm}

\begin{table*}[h]
  \centering
  \caption{}
  \begin{tabular}{|c|r|r|r|r|r|r|r|r|r|r|}
  \hline
  \multirow{2}{*}{ID} &\multirow{2}{*}{SR} & \multicolumn{3}{c|}{Time for SR (s)}& \multicolumn{6}{c|}{Num of gates} \\ \cline{3-5} \cline{6-11} 
   & & Ours & (Shi.1) & (MM.) & Ours & (Shi.1) & (MM.) & (Shi.2 2) & (Shi.2 3) & (Shi.2 4)\\
  \hline
   1  & 1.83 & 3161 & 3994 & 4709 & 3737 & 5536 &4433 & 48103 & 57079 & 57511 \\
   2  & 7.87 & 7142 & 9718 & 10626 & 1595 & 4039 &2300 & 46972 &57511 &57505\\
   3  & 3.79 & 3788 & 5050 & 5425 & 2384 & 4129 &3146  & 51952& 56611& 54883\\
   4  & 2.9 & 3970 & 4781 & 7655 & 3464 & 5128 &5300 &37474& 57487& 57493 \\
   5  & 4.94 & 5035 & 5902 & 7353 & 3235 & 3994 &4249 & 43501 &53971 &57511\\
   6  & 3.53 & 5229 & 7402 & 9731 & 3479 & 6049 &4709 &  47083& 55021& 57511\\
   7  & 2.84 & 4714 & 5839 & 7685 & 4963 & 5824 &6613 & 47227& 57511 &57511\\
   8  & 6.08 & 6559 & 8740 & 9334 & 3227 & 5233 &3902 & 38983& 57487 &57511\\
   9  & 4.01 & 5774 & 7323 & 8633 & 3659 & 5500 &4577 & 47767& 57487 &57511\\
  10  & 2.82 & 3711 & 5373 & 5393 & 2520 & 5497 & 3600 &  42439& 57511 &57511\\
  \hline
  \end{tabular}
  \label{tab3211}
  \end{table*}
In table~\ref{tab3211}, (Shi.1) circuits are generated using the Theorem 1 proposed in~\cite{ShiSDC20}, while (MM.) circuits are generated using the method described in~\cite{mansour2002}, and (Shi.2 $p$) circuits are generated using the Theorem 2 proposed in~\cite{ShiSDC20} under the number of digits of precision $p=2,3,4$.
As observed, the number of gates of (Shi.2 $2,3,4$) are significantly larger than the others, it also consumes more time during SR validation (over 10 hours).

\subsection{Analysis}
We train a standard CNN and 
show 99.73\% accuracy on the test set. 
Then we replace the sigmoid function with the step activation function, 
and the accuracy of the CNN dropped to 99.22\%.
To represent BCNN, compare to BDD and OBDD, 
our algorithm can generate smaller \our{} while keeping the same recognition accuracy. 
It provides certain advantages in later work. 
Note that since~\cite{ShiSDC20}'s strategy is to randomly select multiple orders to generate many OBDDs and select one with the smallest size. 
Although we set the number of random orders to 100, the size of OBDD is still greater than \our.

\begin{figure*}[h]
  \begin{center}
  \centering
  \centering
  \includegraphics[scale=0.38]{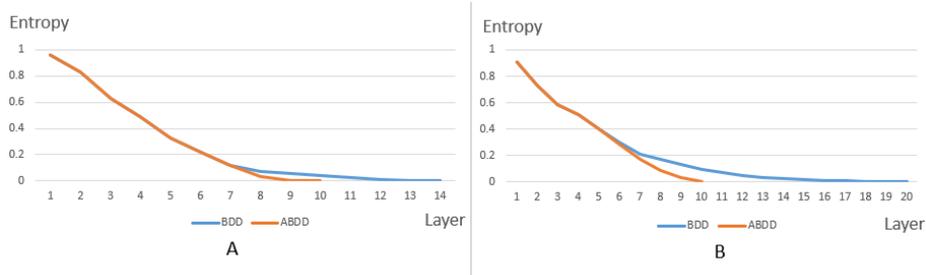} 
  \caption{Illustration of the entropy change of \our{} and BDD at each depth.
  We show the change in the total entropy in each depth of the DD generated by the linear threshold functions of the $3\times 3$ convolution layer (A) and the fully connected layer (B) when using the \our{} and BDD algorithms, respectively.}
  \label{entropy_change}
  \end{center}
  \end{figure*}

Comparing these methods in Table~\ref{tab1}, we observe that the BDD algorithm proposed by Mansour and McAllester~\cite{mansour2002} aims to identify the hypothesis $h$ that yields the most significant reduction in entropy at each node. This approach initially appears to generate a smaller DD due to the intuitive advantage of entropy reduction. However, our observations reveal that this advantage diminishes when the number of samples within a node changes during the merging process. Specifically, if the number of samples remains unchanged, i.e., the entropy of the child node does not decrease to zero, the entropy of both BDD and our algorithm (\our{}) at the same depth remains constant.
However, once this phenomenon occurs, the entropy of \our{} decreases at a faster rate compared to that of BDD, as depicted in Figure~\ref{entropy_change}.
We attribute this phenomenon to the fact that our approach ensures that nodes at the same depth share the same variables, thereby granting us an advantage when merging based on the energy of each node.
When comparing \our{} with the OBDD~\cite{ShiSDC20}, we contend that \our{} offers a precise algorithm for finding the DD with the smallest size, instead of relying on multiple randomizations.
In contrast, when~\cite{ShiSDC20} applies OBDD based on neural unit weights, the time complexity increases exponentially with the size of the weights. In contrast, our algorithm exhibits minimal time consumption, unaffected by changes in weight size. Undoubtedly, this represents a notable advantage of our approach.

\section{Conclusion}

This paper introduced a Boosting-aided method that generates DD with smaller size and time complexities than conventional works. 
Our proposed method's resulting diagram is named ABDD, a variant of standard DD, in which the same variable labels the nodes at the same depth, and the depth is not limited to the number of dimensions of the variable.
Experimental results show that ABDD can be connected in various forms to express NN and can be used to implement various verification tasks efficiently.
Since our method can be used to generate a smaller equivalent circuit of the NN, it can be applied in tasks such as hardware-based transformations of NNs. In the future, we aim to extend our method to more complex NNs.




\bibliographystyle{abbrvnat}
\bibliography{krb}  
\clearpage

\section{Appendix}

\subsection{Proof of Proposition~\ref{propp0}}
\begin{proof} 
By using the property that $\min(q,1-q)\leq G(q)$, 
$\Pr_U\{h_T(x) \neq f(x)\}=\sum_{u\in N(T)}p_u \min(q_u,1-q_u) \leqslant \sum_{u\in N(T)}p_uG(q_u)$.
\end{proof} 

\subsection{Proof of Lemma~\ref{lemma0}}

\begin{proof}
For $(x,y) \in S_u$, we define the balanced distribution of node $u$ as follows:
\begin{equation}\label{eq12}
\begin{split}
d_{(x,y)}^u=
\begin{cases} 
\frac{1}{2\times |\{(x^\prime,y^\prime)\in S_u|y=1\}|},  & \mbox{if }y=1  ~\&  ~(x,y)\in S_u\\
\frac{1}{2\times |\{(x^\prime,y^\prime)\in S_u|y=-1\}|}, & \mbox{if }y=-1 ~\&  ~(x,y)\in S_u\\
0, & \mbox{if } (x,y)\notin S_u
\end{cases}
\end{split}
\end{equation}

Under the balanced distribution $d^{u}$, let 
\begin{equation}\label{eq11}
\begin{split}
\gamma_{u,h_k}=edge_{d^u,f}(h_k)
\end{split}
\end{equation}
be the edge of $h_k$ with respect to $d^u$ and $f$. 
By Lemma~\ref{lemma1-1}, we have
\begin{equation}\label{eq4}
\begin{split}
H_{d^u}(f|h)\leqslant (1-\gamma^2_{u,h}/2)H_{d^u}(f).
\end{split}
\end{equation}


\begin{equation}\label{eq5}
\begin{split}
H_{d^u}(Y|h)=H_{d}(Y|h,\ell_k(x)=u)\leqslant (1-\frac{\gamma^2_{u,h}}{2})H_{d^u}(Y)
\end{split}
\end{equation}



Thus, 
\begin{equation}\label{eq5-2}
\begin{split}
H_U(f|h_k,T_k)
=&\sum_{u \in N(T_k')}p_u G(q_u)\\
=&\sum_{u \in N(T_k)}H_{d^u}(f|h_k)\\
\leq& \sum_{u\in N(T_k)}p_u(1-\gamma^2_{u,h_k}/2)H_{d^u}(f)\\
=&\sum_{u\in N(T_k)}p_u H_{d^u}(f)\\
    & -\sum_{u\in N(T_k)}p_u \frac{\gamma_{u,h}^2}{2}H_{d^u}(f)\\
=&H_U(f|T_k)-\sum_{u\in N(T_k)}p^\prime_u \frac{\gamma_{u,h_k}^2}{2}H_{U}(f|T_k),
\end{split}
\end{equation}
where 
$p^\prime_u =p_u \frac{H_{d_u}(f)}{H_U(f|T_k)}
=p_u\frac{G(q_u)}{\sum_{u\in N(T_k)}p_u G(q_u)}$ 
which is specified in line $3$ of Algorithm~\ref{algsplit}.

Here, we define a distribution $\hat{d}$ for samples in node $u$.
\begin{equation}\label{eq13-1}
\begin{split}
\hat{d}_{(x,y)}=\sum_{u\in N(T_k)} p^\prime_u d_{(x,y)}^u
\end{split}
\end{equation}

By using (\ref{eq5-2}), (\ref{eq13-1}), $\gamma$-WHA, and Jensen's inequality, 
we get
\begin{equation}\label{eq14}
\begin{split}
\sum_{u\in N(T_k)} p^\prime_u\gamma_{u,h_k}^2 
&=\sum_{u\in N(T_k)} p_u^\prime {(\sum_{(x,y)\in S_u} d_{(x,y)}^u y h_k(x))}^2\\
&\geqslant{(\sum_{u\in N(T_k)} p_u^\prime \sum_{(x,y)\in S_u} d_{(x,y)}^u y h_k(x))}^2\\
&={(\sum_{u\in N(T_k)} \sum_{(x,y)\in S_u} p_u^\prime d_{(x,y)}^u y h_k(x))}^2\\
&={(\sum_{(x,y)\in S} \hat{d}_{(x,y)} y h_k(x))}^2\\
&=(edge_{\hat{d},f}(h_k))^2\\
&\geqslant{\gamma}^2.
\end{split}
\end{equation}

By combining \eqref{eq5-2} and \eqref{eq13-1}, 
\eqref{eq14} can be rewritten as 
\begin{equation}\label{eq5-3}
\begin{split}
H_U(f|h_k,\ell_k)
&\leqslant (1-\gamma^{2})H_U(f|T_k)
\end{split}
\end{equation}
as desired.
\end{proof}

\subsection{Proof of Theorem~\ref{mainth}}
\begin{proof}
By definition of $\hat{\lambda}$ in Algorithm~\ref{alg1}, 
we have $H_U(f|T_k,h_k)=(1-\hat{\lambda})H_U(f|T_k)$ at each iteration $k$.
Therefore, by Lemma~\ref{lemma-merge}, 
$H_U(f|T_{k+1})\leq (1-\hat{\lambda}/{2})H_U(f|T_k)$. Furthermore, $\hat{\lambda}$ is always larger than 
$\gamma^2/2$, implying that 
$H_U(f|T_{k+1})\leq (1-\gamma^2/{4})H_U(f|T_k)$.
Then, after $K$ iterations, we have 
\begin{align*}
    H_U(f|T_K) 
    \leq \left(1-\frac{\gamma^2}{4}\right)^K \leq e^{-\frac{K\gamma^2}{4}},   
\end{align*}
as $H_U(f)\leq 1$. 
Thus, $K=O((1/\gamma^2)\ln(1/\varepsilon))$ iterations suffices to ensure that 
$\Pr_U\{h_T(x)\neq f(x)\}\leq \varepsilon$.
Since the width of the ABDD $T$ is 
$O((1/\gamma^2)(\ln(1/\gamma)+\ln(1/\varepsilon)))$ by Lemma~\ref{lemma-merge} and 
its depth is $K$, the size is bounded by 
$O((\ln (1/\varepsilon) /\gamma^4) (\ln(1/\varepsilon) +\ln (1/\gamma) ))$, as claimed.
\end{proof}

\subsection{Proof of Lemma~\ref{mainlemma}}

\begin{proof}
Let $\rho$ be the optimal solution of (\ref{eq:margin}).
By duality, $\rho$ is identical to the optimal
solution $\gamma$ of the dual problem of (\ref{eq:margin}):
\begin{align*}
    & \min_{d,\gamma} \gamma \\
\text{s.t. \quad} & 
\sum_{x \in \{-1,1\}^n} d_x f(x)h_i(x) \leq \gamma
\text{ for any $i \in [2n]$,} \\
& d \geq 0, \\
& \sum_{x: f(x) = 1} d_x = \sum_{x: f(x) = -1} d_x = 1/2. \\
\end{align*}
Apparently, the optimal $\gamma$ should satisfy
$\gamma = \max_{h \in \calH} \sum_x d_x f(x)h(x)$,
which implies the lemma.
\end{proof}

\end{document}